\def\year{2020}\relax
\documentclass[letterpaper]{article} % DO NOT CHANGE THIS
\usepackage{aaai20}  % DO NOT CHANGE THIS
\usepackage{times}  % DO NOT CHANGE THIS
\usepackage{helvet} % DO NOT CHANGE THIS
\usepackage{courier}  % DO NOT CHANGE THIS
\usepackage[hyphens]{url}  % DO NOT CHANGE THIS
\usepackage{graphicx} % DO NOT CHANGE THIS
\usepackage{graphicx}  % DO NOT CHANGE THIS
\usepackage{subcaption}
\usepackage{ctable}
\usepackage{bbold}
\usepackage{amsmath}
\usepackage{float}

\graphicspath{{figures/}}

\newcommand*{\Eqref}[1]{Eq.~\eqref{#1}}
\def\*#1{\mathbf{#1}}

\title{Recurrent Point Processes for Dynamic Review Models}
\author{Kostadin Cvejoski,\textsuperscript{\rm1}
Rams\'es J. S\'anchez,\textsuperscript{\rm1}
 Bogdan Georgiev,\textsuperscript{\rm1}\\
 \bf \Large Jannis Schuecker, \textsuperscript{\rm3}
 \bf \Large Christian Bauckhage\textsuperscript{\rm3},
 C\'esar Ojeda,\textsuperscript{\rm4}\\
 \textsuperscript{1}Competence Center Machine Learning Rhine-Ruhr\\
 \textsuperscript{2}B-IT, University of Bonn, Bonn, Germany\\
 \textsuperscript{3}Fraunhofer Center for Machine Learning and Fraunhofer IAIS, 53757 Sankt
 Augustin, Germany\\
 \textsuperscript{4}Berlin Center for Machine Learning and TU Berlin, 10587 Berlin, Germany\\
 \{kostadin.cvejoski, bogdan.georgiev, jannis.schuecker, christian.bauckhage\}@iais.fraunhofer.de,\\ojeda.marin@tu-berlin.de, sanchez@bit.uni-bonn.de
 }
\begin{document}

\maketitle

\begin{abstract}
Recent progress in recommender system research has shown the importance of including temporal representations to improve interpretability and performance. Here, we incorporate temporal representations in continuous time via recurrent point process for a dynamical model of reviews. Our goal is to characterize how changes in perception, user interest and seasonal effects affect review text.
\end{abstract}

\section{Introduction}
Costumer reviews provide a rich and natural source of unstructured data which can be leverage to improve interactive and conversational recommender system performance \cite{liu2019daml}. Reviews are effectively a form of recommendation. Although causal and temporal relations have been know to improve the performance of recommender systems \cite{wu2017recurrent}, recent natural language process (NLP) methodologies for rating and reviews \cite{zheng2017joint} lack behind at incorporating temporal structure in language representations. In the present work, we exploit recurrent neural network (RNN) models for point process and include neural representations of text to characterize costumer reviews. Our goal is to capture the changes in taste and importance of items during time, and how such changes reflect on the text produced by the different users.

The reviews research have sought to characterize usefulness and generation of reviews \cite{fan2019product,novgorodov2019generating} and provide better representations for rating prediction \cite{esmaeili2019structured}. The need to interact with costumers have lead to question answering solutions \cite{chen2019driven,yu2018aware}. Deep neural networks models for rating predictions use embedding representations as well as convolutions neural networks \cite{catherine2017transnets}.  
Dynamic models of text however have shown more success from the bayesian perspective within topic models \cite{rudolph2018dynamic,wang2012continuous}. Self exciting point processes have allow for clustering of document streams \cite{du2015dirichlet,he2015hawkestopic}. Different from these works, we focus on the temporal aspects of the text for each review.
\begin{figure}[h]
     \centering
      \begin{subfigure}[h]{0.185\textwidth}
         \centering
         \includegraphics[width=\linewidth]{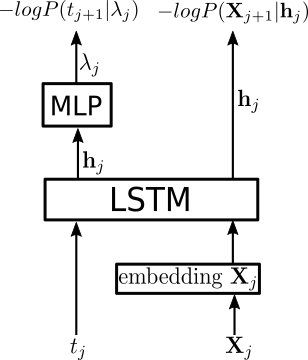}
         \caption{Auto-regressive model (LSTM-BoW)}
         \label{fig:lstm_bow}
     \end{subfigure}
     \hspace{8mm}
      \begin{subfigure}[h]{0.18\textwidth}
         \centering
         \includegraphics[width=\linewidth]{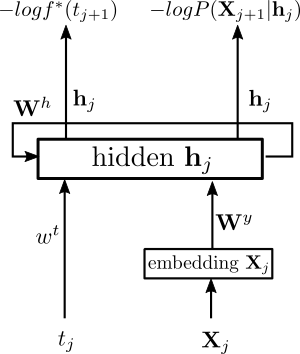}
         \caption{Recurrent Point Review Model (RPRM)}
         \label{fig:rmtpp_bow}
     \end{subfigure}
    \caption{Bag of words models used for predicting the next content and the time of the next review.}
    \label{fig:models}
\end{figure}
\section{Recurrent Point Review Model (RPRM)}
Consider an item  $a$ (e.g. a business, service or movie) and assume that, since its opening to the public, it has received a collection of $N_a$ reviews $\*r^a = \{(\*x^a_{j},t^a_j)\}_{j=1}^{N_a}$, where $t^a_j$ labels the creation time of review $\*r^a_j$ and $\*x^a_j = (\*w^j_1,...,\*w^j_{M_j})$ corresponds to its text\footnote{In what follows we shall drop the index $a$ over items.}. Such a collection of reviews effectively defines a point process in time.
Our main idea is to model these point processes as RNNs \textit{in continuous time} and use their hidden representations, which encode the nonlinear relations between text and timing of past reviews, to predict how the reviews' text of a given item changes with time.
The model thus consists of two components: a point process model which leverages the information encoded in the review text and a dynamic neural text model. 
% which uses the point process history.

We start by transforming the text of each review into a \textit{bag of words} (BoW) representation $\mathbf{X}_j\in \mathbb{R}^V$, where $V$ is the vocabulary size \cite{NIPS2009_3856}.
% Since its opening to the public, a given item $a$ (e.g. a business, service or movie) receives a collection of reviews $\*r^a = \{(\*x^a_{j},t_j)\}_{j=1}^{N_a}$, where $t_j$ corresponds to the creation time of review $\*r^a_j$ and $\*x^j$ to its text.
%
% For each item $a$ our data set contains a set of $N_a$ reviews  $\*r^a = (\*x^a_{j},t_j)$ where $t_j$ corresponds to the creation time of review $\*r^a_j$ and $\*x^j = (\*w^j_1,...\*w^j_M)$ its corresponding text. We transform the text into a \textit{bag of words} (BoW) representation $\mathbf{X^j}$, where we follow the procedure of \cite{NIPS2009_3856}.
%
%%% Let $\*X^n$ be the \textbf{Bag of words} representation of the documents and let $\*x^n_i$ be the one hot representation of the word at position $x^n_i$ for document $n$ occurring at time $t_n$. In our case, the encoder (in this case deterministic i.e. no Gaussian), creates the representation for the data.  
%
%%% shopping_business_rnn_model_embedding_size_16_model_cell_type_hidden_size_16_data_loader_batch_size_128_data_loader_bptt_len_20_data_loader_t_max_1_optimizer_lr_0.001/0911_154150 MSE 8995 Perplexity exp(6.32)
%
%%% hopping_business_bow_model_embedding_size_16_model_cell_type_hidden_size_64_data_loader_batch_size_128_data_loader_bptt_len_20_data_loader_t_max_100.0_optimizer_lr_0.001/0811_223303 MSE 8893 Perplexity(6.259)

\textbf{Recurrent Point Process (RPP):}
\label{ssec:RMPP}
% Now we write the likelihood of a point process induced by an intensity function $\lambda^{*}(t)$, and defined through a RNN, following \cite{NeuralHawkes}, \cite{RecurrentTemporal}.
Let us consider a point process with compact support $\mathcal{S} \subset \mathbb{R}$. Formally, we write the likelihood of a new arrival (i.e. a new review) $\*r_{j+1}$ as an inhomogeneous Poisson process between reviews, conditioned on the history $\mathcal{H}_j \equiv  \{\*r_1,...,\*r_j\}$\footnote{a.k.a. filtration.} \cite{VereJones}. 
For the one-dimensional processes we are concerned here, the conditional likelihood function reads
\begin{equation}
f^*(t) = \lambda^*(t) \exp\left\{ \int^{t}_{t_j}\lambda^*\left(t'\right)dt' \right\}\, ,
\label{eq:point_likelihood}
\end{equation}
where $\lambda^*$ is (locally) integrable and is known as the intensity function of the point process.
Following \cite{NeuralHawkes}, \cite{RecurrentTemporal}, we define the functional dependence of the intensity function to be given by a RNN with hidden state $\*h_j \in \mathbb{R}^d$, where an exponential function guarantees that the intensity is \textit{non-negative}
\begin{equation}
\lambda^*(t) = \exp{\left\{\*v^t\cdot\*h_j + w^t\left(t-t_j\right) + b^t\right\}} \label{eq:intensity}\, .
\end{equation}
Here the vector $\*v^t \in \mathbb{R}^d$ and the scalars $w^t$ and $b^t$ are trainable variables. The update equation for the hidden variables of the recurrent network can be written as a general non-linear function 
\begin{equation}
\*h_j = f_{\theta}(t_{j}, \*X_j, \*h_{j-1}),
\label{eq:transitionfunction}
\end{equation}
% $\*h_j = f_{\theta}(t_{j}, \*X_j, \*h_{j-1})$,
where $t_j$ and $\*X_j$ label the creation time and the text's BoW representation of review $\*r_j$, respectively, and $\theta$ denotes the network's parameters. We thus use the BoW representation of the review text as marks in the recurrent marked temporal point process \cite{du2016recurrent}. Inserting \Eqref{eq:intensity} into \Eqref{eq:point_likelihood} and integrating over time immediately yields the likelihood $f^*$ as a function of $\*h_j$. 
% Performing the integration in \Eqref{eq:point_likelihood} one obtains
% \begin{multline}
% f^*(t) =  \exp\left\{\*v^t \cdot \*h_j + w^t\left(t-t_j\right) \right.\\ \left.+ b^t + \frac{1}{w^t}\exp\left\{\*v^t\cdot \*h_j + b^t\right\} \right. \\ 
%  \left. - \frac{1}{w^t}\exp\{\*v^t\cdot\*h_j + w^t\left(t-t_j\right) + b^t\} \right\}.
% \end{multline}

% We can now learn the parameters of the point process model by maximizing the model log-likelihood $\mathcal{L}_{\mbox{RPP}} = \sum_{j}\log f^*(\delta_{j+1}|\*h_j)$, where $\delta_{j+1} = t_{j+1}-t_j$ denotes the inter-arrival time.

\textbf{Dynamic Neural Text Model:} To model the text component of reviews we assume the words in review $\*r_{j+1}$ are generated independently, conditioned on $\*h_j$, the temporal representation of the RPP above. Specifically, we follow \cite{miao2016neural} and write the conditional probability of generating the $i$th word $\*w^{j+1}_i$ of the $(j+1)$th review as
% The modeling of the text of each review is done using the idea of neural variational document model \cite{miao2016neural}. In our case however, we do not use variational methods. Let $\*h \in \mathbb{R}^V$ be a continuous hidden variable which generates all the words independently. We encode the bag of word representations $\mathbf{X}^N$ into $\mathbf{h}_j$ through projecting the BoW into the lower dimensional space. The generation model $p(\*X^j|\*h_j)$ is defined as,
\begin{align}
    p_\theta(\*w^{j+1}_i|\*h_j) =&  \frac{\exp{\{-z(\*w^j_i,\*h_j)\}}}{\sum_{k=1}^V \exp{\{-z(\*w^j_k,\*h_j)\}}},  \label{eq:word_prob} \\ 
    z(\*w^j_i,\*h_j) =& -\*h^T_j \*R \, \*w^j_i - \*b \, \*w^j_i, 
\end{align}
where $\*R \in \mathbb{R}^{d \times V}$ and $\*b \in \mathbb{R}^V$ are trainable parameters, and $\*w^{j}_i$ is the one-hot representation of the word at position $i$. 
The complete log-likelihood of the RPRM model can then by written as
\begin{equation}
    \mathcal{L}=\sum_{a=1}^{N}\sum_{j=1}^{N_a}(\log f^*(\delta^a_{j+1}|\*h_j)+ \log P(\mathbf{X}^a_{j+1}| \*h_j)),
\end{equation}
where $\delta^a_{j+1} = t^a_{j+1}-t^a_j$ denotes the inter-review time for item $a$ and $P(\mathbf{X}^a_{j+1}| \*h_j)$ is a multinomial distribution over word probabilities (\Eqref{eq:word_prob}) and counts.

\textbf{Baseline model:} In order to test our model we define LSTM-BoW, which models the inter-review time $\delta_{j+1}$ as the mean of an exponential distribution with parameter $\lambda_{\phi}(\*h'_j)$, and the probability over words $p(\*w^{j+1}_i|\*h'_j)$ as $\mbox{softmax}(g_{\phi}(\*h'_j))$. The functions $\lambda_{\phi}$ and $g_{\phi}$ are given by neural networks with parameter $\phi$, and $\*h'_j = f_{\phi}(t_{j}, \*X_j, \*h'_{j-1})$ is the hidden state of an LSTM network \cite{LSTM}.
% For a base line model LSTM-BoW Fig. \ref{fig:lstm_bow} we use an recurrent neural network (LSTM) as regression. First, we embed the bag of words vector into lower dimension space, then this embedding is concatenated with the time of the review and passed to the LSTM. 
% The model is trained using maximal likelihood. 
We also consider additional LSTM and RPP models which only take $t_j$ as input, as to check whether the BoW representation $\*X_i$ helps in the prediction of the inter-review times.

\section{Experiments and Results}
We test our models on the Yelp19 dataset\footnote{https://www.yelp.com/dataset}. Specifically we take all reviews for businesses that are labeled with the \textit{shopping} parent category from \textit{01 Jan 2016} to \textit{30 Nov 2018}. The creation time of a review is defined as the difference in days between the original timestamp and \textit{01 Jan 2016}. Next, we group reviews by business. All businesses with less than 5 reviews are removed. The text from each review is converted into a BoW vector of size 2000 \cite{NIPS2009_3856}. The result from the preprocessing is a dataset that has in total 262193 reviews, 27185 businesses, 174122 users, 1910299 sentences and 13209813 words. On average we have 9.6 reviews per business with standard deviation of 22.8. Each review has on average 7.2 sentences and 50.2 words, with 5.9 and 46.14 standard deviation respectively. In the experiments, we randomly split each dataset into two parts: training set (80\%) and test set (20\%). We use grid search for hyper-parameters finding. 
% The Bow embedding size is tuned from [16, 64, 128, 256], the hidden state size is tuned from [16, 64, 256]. The batch size is tuned from [32, 64, 256] and the backpropagation trough time window size is tuned from [10, 20, 30]. We use Adam for optimizer with weight decay 0.001 and learning rate tuned from [0.001, 0.0001, 0.00001]. The methods are implemented using the PyTorch v1.3\footnote{https://pytorch.org/}.

We trained all models on maximum likelihood and use two evaluation metrics: Root-mean-squared error (RMSE) on the inter-review times and predictive perplexity on the review text. The latter is defined as $\mathcal{PP}= \exp\left\{-\frac{1}{T}\sum^T_{j=1}\frac{1}{|\mathcal{H}_j|}\sum_{i\in \mathbf{x}_j}\frac{\log p(\mathbf{w}^j_i|\*h_{j-1})}{M_j}\right\}$ \cite{wang2012continuous}, where $M_j$ is the number of words in review $\*r_j$ and $|\mathcal{H}_j|$ is the number of reviews at time $t_j$. Our results are presented in Table \ref{tab:results} and show that the best model in both metrics is the Recurrent Point Review Model. Note also that the models that leverage the information encoded in the text (through $\*X_j$) show improvement of the RMSE (with respect to the inter-review time) over the models which do not see $\*X_j$. 
% What is interesting to notice is that the RPRM model performs better than the LSTM+BoW model in terms of predictive perplexity.

% \begin{table*}[h!]
%     \centering
%     \begin{tabular}{ cccccc } 
%         \textbf{Dataset} & \textbf{\#reviews} & \textbf{\#items} & \textbf{\#users} & \textbf{\#sentences} & \textbf{\#words} \\ 
%         \specialrule{.1em}{.05em}{.05em} 
%         Yelp-Shopping & 262 193 & 27 185 & 174 122 & 1 910 299 &  13 209 813 \\ 
%     \hline
%     \end{tabular}
%      \caption{Yelp shopping category dataset statistics.}
%     \label{tab:dataset_stats}
% \end{table*}

% \begin{table}[h]
%     \centering
%     \begin{tabular}{ lrrrrrrrr } 
%         &  \textbf{mean} & \textbf{std} &  \textbf{max} & \textbf{50\%}  & \textbf{95\%} & \textbf{99\%} \\ 
%       \specialrule{.1em}{.05em}{.05em} 
%       reviews & 9.6 & 22.8 & 1 153 & 4 & 33 & 84 \\ 
%       sentences & 7.2 & 5.9 & 312  & 6 & 18 & 31 \\ 
%       words & 50.3 & 46.14 & 624  & 4 & 136 & 237 \\ 
%     \hline
%     \end{tabular}
%      \caption{Number of reviews per business statistics.}
%     \label{tab:arrivals_stats}
% \end{table}

\begin{table}[t]
    \centering
    \begin{tabular}{ lrrr } 
        \textbf{Model}  & \textbf{RMSE} & $\mathbf{R^2}$ & \textbf{Pred. Perplexity}\\ 
        \specialrule{.1em}{.05em}{.05em} 
        LSTM & 96.8813  & 0.1788 & - \\
        RPP & 96.3794 & 0.1873 & - \\
        LSTM-BoW & 95.3414 & 0.2046 & 519.90\\ 
        RPRM & \textbf{92.3850} & \textbf{0.2533} & \textbf{511.32}  \\ %511.322  63728
    \hline
    \end{tabular}
     \caption{Model performance on RMSE, $R^2$ and predictive perplexity. 
    %  The RMSE and $R^2$ are on the time component, the predictive perplexity is on the BoW representation.
     }
    \label{tab:results}
\end{table}
%%%\textbf{Predictive Perplexity}
%%%Following 
%\begin{equation}
%\mathcal{P}(t) = \exp\left\{-\frac{1}{D_t}\sum_{d\in D_t}\frac{\log %p(w_d|D_{1:t-1})}{N_d}\right\}
%\end{equation}
\section{Conclusion and Future Work}
In this work we incorporate a bag of word language model as the marks of a recurrent temporal point process. This creates a model which characterize temporal and causal representation for text, allowing for a richer representation for costumers reviews. We show that this improves predictive performance for the time of the reviews, as well as opening the door for \textit{text prediction}. We will extend this methodology for rating prediction as well as more complex models of text.\footnote{One part of this research has been funded by the Federal Ministry of Education and Research of Germany as part of the competence center for machine learning ML2R (01|S18038A).}
\bibliographystyle{aaai}
\bibliography{text_point_aaai}
\end{document}